\documentclass[10pt,conference]{IEEEtran}
\IEEEoverridecommandlockouts

\usepackage{cite}
\usepackage[font=small,labelfont=bf]{caption} 
\ifCLASSINFOpdf
   \usepackage[pdftex]{graphicx}
   \graphicspath{{figs/}}
   \DeclareGraphicsExtensions{.pdf,.jpeg,.png}
\else
   \usepackage[dvips]{graphicx}
   \graphicspath{{../figs/}}
   \DeclareGraphicsExtensions{.eps}
\fi
\usepackage[cmex10]{amsmath}
\usepackage{amsthm}

\usepackage{algorithmic}

\usepackage{array}

\ifCLASSOPTIONcompsoc
  \usepackage[caption=false,font=normalsize,labelfont=sf,textfont=sf]{subfig}
\else
  \usepackage[caption=false,font=footnotesize]{subfig}
\fi
\usepackage{url}

\newif\iffinal
\finaltrue
\newcommand{\cmtid}{205}

\iffinal
\else
\usepackage[switch]{lineno}
\fi

\usepackage{color}                      
\definecolor{vcolor}{rgb}{0.75,0.5,0.0} 
\definecolor{dgreen}{rgb}{0.1, 0.8, 0.3}  

\usepackage{tabu}                      
\usepackage{booktabs}                  
\usepackage{amsfonts}                  
\usepackage[acronym]{glossaries}       
\makeglossaries
\glsdisablehyper                       

\usepackage{comment}                   
\usepackage{subfig}                    
\newcommand{\figwidth}{0.32}

\newcommand{\VM}[1]{\textcolor{magenta}{VM: #1}}

\newcommand{\XP}[1]{\textcolor{cyan}{XP: #1}}
\newcommand{\myparagraph}[1]{\smallskip\noindent\textbf{#1}\,\,}

\newacronym{gae}{GAE}{Graph AutoEncoder}
\newacronym{ml}{ML}{Machine Learning}
\newacronym{dtw}{DTW}{Dynamic Time Warping}
\newacronym{tsne}{t-SNE}{t-distributed Stochastic Neighbor Embedding}
\newacronym[plural=POIs,longplural={Points of Interest}]{poi}{POI}{Point of Interest}
\newacronym{uva}{UVA}{Urban Visual Analytics}
\newacronym{va}{VA}{Visual Analytics}

\newcommand{\kmeans}{\emph{k}-means}

\IEEEoverridecommandlockouts
\IEEEpubid{\makebox[\columnwidth]{979-8-3315-8951-6/25/\$31.00~\copyright2025 IEEE \hfill}
\hspace{\columnsep}\makebox[\columnwidth]{ }}

\begin{document}
%
\title{Exploring Urban Factors with Autoencoders:\\Relationship Between Static and Dynamic Features}


\iffinal




%
\author{\IEEEauthorblockN{Ximena Pocco \IEEEauthorrefmark{1},
Waqar Hassan\IEEEauthorrefmark{1},
Karelia Salinas\IEEEauthorrefmark{1},
Vladimir Molchanov\IEEEauthorrefmark{2} and
Luis G. Nonato\IEEEauthorrefmark{1}}
\IEEEauthorblockA{\IEEEauthorrefmark{1}
ICMC, University of Sao Paulo, Sao Carlos, Brazil} \IEEEauthorblockA{\IEEEauthorrefmark{2} Münster University, Westphalia, Germany} 
}

\else
  \author{SIBGRAPI Paper ID: \cmtid \\ }
  \linenumbers
\fi

\maketitle
\begin{abstract}
Urban analytics utilizes extensive datasets with diverse urban information to simulate, predict trends, and uncover complex patterns within cities. While these data enables advanced analysis, it also presents challenges due to its granularity, heterogeneity, and multimodality. To address these challenges, visual analytics tools have been developed to support the exploration of latent representations of fused heterogeneous and multimodal data, discretized at a street-level of detail. However, visualization-assisted tools seldom explore 
the extent to which fused data can offer deeper insights than examining each data source independently within an integrated visualization framework. In this work, we developed a visualization-assisted framework to analyze whether fused latent data representations are more effective than separate representations in uncovering patterns from dynamic and static urban data. The analysis reveals that combined latent representations produce more structured patterns, while separate ones are useful in particular cases.

\end{abstract}



\IEEEpeerreviewmaketitle

\vspace*{-0.56mm}
\section{Introduction}
Urban analytics harnesses large, diverse datasets to simulate, forecast, and detect patterns in cities~\cite{batty2019urban}. However, these datasets often vary in granularity and structure. Granularity refers to the spatial scale of data, e.g., socioeconomic data is aggregated by census tract, while public amenities are geolocated at the point level. 
Moreover, urban data typically falls into two main categories: static and dynamic. Static data, such as infrastructure and demographics, changes slowly over time and is typically tabular. Dynamic data, e.g., crime reports or air quality, varies frequently and is captured as a time series. As static factors can influence dynamic phenomena, both must be analyzed together to better understand the complexity of urban phenomena.

Street-level discretization effectively handles data heterogeneity and granularity, enabling fine-grained modeling across diverse applications~\cite{Deng24}.
In this context, \gls{ml} models that generate latent representations of geolocated data are widely used~\cite{liu2020urban,hassan2024modeling,zou2025deep}, supporting tasks like neighborhood similarity analysis~\cite{jin2024learning}, \gls{poi} recommendation~\cite{li2021discovering}, and anomaly detection~\cite{zhang2020urban}.


\gls{va} tools have been developed to explore latent representations~\cite{deng2023survey,garcia2020visualization,yang2024exploring}, targeting static~\cite{lee2024latent}, dynamic~\cite{garcia2021cripav}, or fused data~\cite{moreira2024curio}. However, few studies examine whether fusing static and dynamic data reveals more insights than analyzing them separately, as some patterns may be unique to either fused or individual views. Moreover, visualization tools have not been properly exploited to assist in the analysis and comparison of different data fusion mechanisms.

This work fills this gap by proposing a visualization-assisted methodology to analyze whether fused data representations offer more insight than using static or dynamic data alone. We present a methodology using graph autoencoders~\cite{majumdar2018graph} to compare different models designed to learn fused and separate latent representations of multimodal data discretized at a street-level granularity. Our interactive visual tool combines linked scatter plots with coordinated views to support the interpretation of the different data fusion schemes.

Through experiments on both synthetic and real-world data, we demonstrate that combining static and dynamic features can yield richer insights. Specifically, the synthetic data experiments quantitatively and qualitatively highlight the effectiveness of data fusion in jointly representing static and dynamic information. Moreover, case studies with real data demonstrate that fused representations improve the understanding of urban phenomena, revealing that the proposed data fusion models tend to place greater emphasis on dynamic data while still accounting for the importance of static information.

In summary, the main contributions of this work are:
\begin{itemize}
    \item A methodology to generate latent representations of fused and individual static/dynamic data at street level.
    \item An experiment involving synthetic data that shows the power of fusion mechanisms in representing static and dynamic data together.
    \item A visualization tool supporting linked exploration of fused and separate data to uncover urban patterns.
    \item Case studies demonstrating when and why fused or separate representations yield important insights.
\end{itemize}

\section{Related Work}

\gls{uva} systems help reveal complex spatiotemporal patterns in cities. Here, we focus on tools for analyzing static and dynamic urban data; see surveys\cite{deng2023survey,feng2022survey,ferreira2024assessing} for a broader overview. 
Most existing systems build upon representation learning mechanisms based on geospatial networks combined with dimensionality reduction methods. A comprehensive discussion about geospatial networks-based representations and dimensionality reduction is beyond the scope of this work.
Interested readers may refer to the surveys~\cite{schottler2021visualizing, nonato2018multidimensional} 
for an in-depth discussion about those topics.

\myparagraph{Static data}
Some \gls{uva} systems focus on static data, i.e., slow-changing information
such as census data or facility locations. For example, Chen et al.~\cite{Chen24} use latent \gls{poi} to explain urban performance metrics. Static geospatial data also supports regionalization analysis, clustering neighborhoods by shared attributes~\cite{Yu23} to promote equitable urban planning~\cite{lyu2023if}. Several tools embed static data into street-network graphs to support traffic analysis~\cite{weng2018srvis,weng2021}, assess \gls{poi} accessibility~\cite{feng2020topology}, and perform multilevel geospatial analysis~\cite{Deng24}.

\myparagraph{Dynamic data}
Dynamic data captures events like traffic, accidents, and crime which might be updated hourly or daily. Visualization tools often use latent representations to manage such data. García-Zanabria et al.~\cite{Garcia20, garcia2021cripav} employed autoencoders to extract patterns from crime time series at the micro-scale. Wang et al.~\cite{wang2013visual} represented traffic jams using spatiotemporal tuples mapped onto road networks. CATOM~\cite{jung2024catom} encodes causal traffic relations in a dynamic matrix.

\myparagraph{Combined data}
Combining static and dynamic data offers deeper insights, especially in domains like crime and transportation. Curio~\cite{moreira2024curio} facilitates collaborative urban analysis by integrating data preparation, management, and visualization. Hou et al.~\cite{hou2022integrated} demonstrated how static socioeconomic data contextualizes dynamic crime patterns.
Zheng et al.~\cite{ZHENG2009484} showed that integrating both data types improves forecasting using neural networks and Bayesian methods. Similarly, Huang et al.~\cite{huang2019deep} proposed a Dynamic Fusion Network for accident prediction. Liang et al.~\cite{liang2022towards} used \gls{ml} to predict hourly crime based on weather, holidays, and history, highlighting the role of temporal and spatial context.


Despite these advances, few has been done towards understanding the behavior of fusion mechanisms, particularly the lack of comparative evaluations on how fusion design influences learned embeddings. While most prior work focuses on predictive outcomes or visual presentation, our work leverages \gls{gae} to fuse static and dynamic data into a unified representation (see Sec.~\ref{Sec: fusion_approaces}). Moreover, we provide a methodology to visually analyze the latent spaces resulting from four distinct fusion strategies, along with a visualization tool to compare and interpret fusion models, thereby uncovering properties of data fusion mechanisms while supporting the analysis of complex multimodal data.

\section{Fusion Strategies and Models} \label{Sec: fusion_approaces}


Spatiotemporal datasets integrating geospatial and time-dependent information pose challenges for representation learning. In our context, both data types are discretized on a spatial street graph of São Paulo, where nodes represent geolocated intersections with static socioeconomic attributes and dynamic monthly crime counts. The objective is to learn compact, informative latent representations for each node that integrate both static and dynamic features.
We employ \gls{gae}s to encode multimodal urban data into high-dimensional embeddings, which are then projected into 2D via \gls{tsne} to support the visualization of clusters and patterns.


\subsection{\gls{gae} as a Representation Learning Framework}

To encode node-level features while preserving spatial structure, we employ a \gls{gae} architecture composed of an \textit{encoder} and a \textit{decoder}. The encoder projects node attributes into a latent space using two stacked GraphSAGE convolutional layers (\texttt{SAGEConv}) with ReLU activations \cite{hamilton2017inductive}, while the decoder mirrors this structure. Unlike standard \glspl{gae}, we do \textbf{not} reconstruct the adjacency matrix; the model is trained solely to reconstruct node features, aligning with our focus on attribute encoding rather than structural inference.


In the fusion models, we integrate a sigmoid-based attention gate to balance static and dynamic node features. This gate assigns weights to each feature dimension, highlighting sparse yet informative dynamic signals. The weighted features are processed by GraphSAGE layers for latent encoding and a GraphSAGE-based decoder for reconstruction. Inspired by self-attention and gating mechanisms in GNNs \cite{velivckovic2017graph}, this design is well-suited for settings where dynamic events, such as spatiotemporal crime patterns, are relatively rare.


A central hyperparameter is the latent space dimensionality, tuned through empirical tests to balance: a) minimizing reconstruction error, favoring higher dimensions, and b) achieving compact, observable representations, favoring lower dimensions. Semantic interpretability remains an open challenge beyond the scope of this work.


\subsection{Attribute Fusion in Spatiotemporal Encoding}
The methodological challenge lies in the \textit{fusion} of heterogeneous node attributes, specifically, the integration of static and dynamic features.
Fusion in this work refers to architectural strategies within \glspl{gae} that integrate static and dynamic data, shaping the latent space and distinguishing our approach from prior surface-level or visual integration methods~\cite{hou2022integrated}.
We systematically investigate different fusion architectures (early, late, and hierarchical fusion) and evaluate their effectiveness.





\begin{figure}[t]
    \centering
    \vspace*{-0.5cm}
    \includegraphics[width=\linewidth]{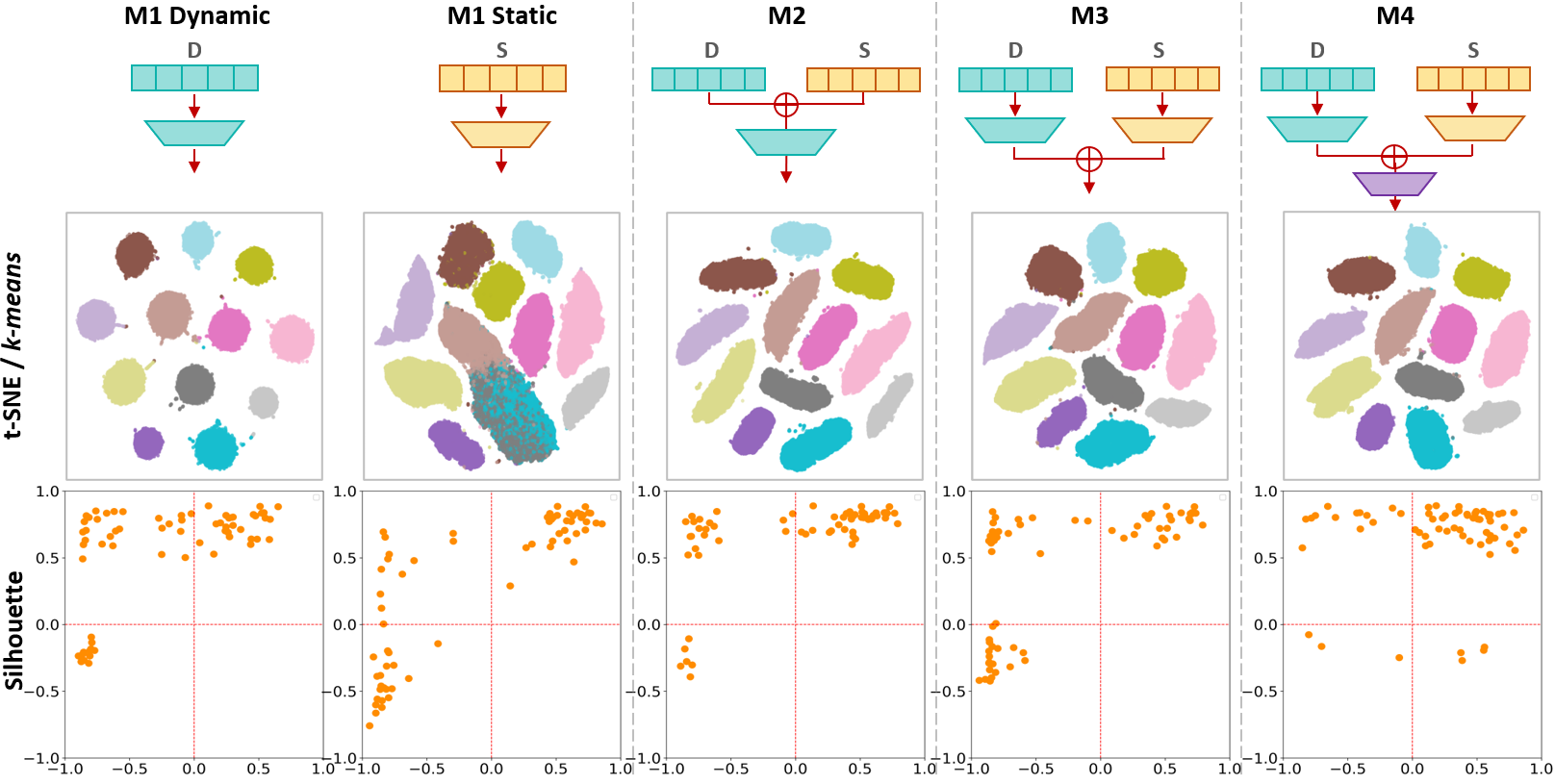} 
    \caption{Schematic illustration of the fusion strategies for combining static and dynamic features in \glspl{gae} (first row). \gls{tsne} projections of the synthetic dataset (middle row) and resulting silhouette pairs of clusters (last row). 
    }
    \label{fig:fusion_models}
    \vspace*{-0.5cm}
\end{figure}

Fig.~\ref{fig:fusion_models} presents a schematic overview of the four proposed models (\textbf{M1}–\textbf{M4}), highlighting their fusion strategies for static (S) and dynamic (D) node features. The middle row shows \gls{tsne} projections of the resulting embeddings with \kmeans\ clusters. The bottom row displays the silhouette plots to evaluate cluster quality. The specific steps and operations involved in each stage are described in the next section. 
\subsection{Fusion-Encoding Models}
\noindent\emph{\textbf{M1} -- Independent Embedding of Static and Dynamic Features}
In this baseline, two independent \glspl{gae} are trained separately:
One processes static socioeconomic features and
the other encodes dynamic crime data.
Each model produces its own latent space. This approach avoids any fusion and treats each modality independently. The two embeddings are analyzed separately in downstream evaluations.

\noindent\emph{\textbf{M2} -- Early Fusion via Feature Concatenation}
Here, static and dynamic features are \textit{concatenated at the input level} to form a single feature vector for each node. This unified representation is passed into a \textit{single \gls{gae}}, which learns a joint embedding. This early fusion strategy forces the model to learn a shared representation across both modalities from the beginning.

\noindent\emph{\textbf{M3} -- Late Fusion of Embeddings}
Two \glspl{gae} are trained independently. However, their embeddings are \textit{concatenated post-training} to form a \textit{composite embedding}.
This strategy assumes that each modality captures complementary information and defers fusion until after the individual latent spaces are learned.

\noindent\emph{\textbf{M4} -- Hierarchical Fusion via Stacked \glspl{gae}}
This model introduces a \textit{multi-stage architecture}:
1) Two initial \glspl{gae} are trained separately on static and dynamic data;
2) Their embeddings are \textit{concatenated} to produce an intermediate fused representation;
3) A third \gls{gae} is trained on this fused embedding to produce a final high-level latent space.
Unlike \textbf{M3}, all three \glspl{gae} are trained \textit{jointly}, enabling end-to-end optimization and layered abstraction of features. 
This hierarchical design aims to better capture complex interactions between static and dynamic signals.

These four architectures reflect progressively deeper integration of heterogeneous data, from fully independent encoding to hierarchical fusion. 
In Sec.~\ref{sec: synthetic_evaluation} and \ref{sec:visual-analysis-tool}, we compare these models in terms of clustering quality
and latent space structure to identify effective spatiotemporal fusion strategies. To ensure reproducibility, all code and data used in this study are publicly available in the GitHub repository
(\url{https://github.com/giva-lab/sib_data_fusion})


\section{Data Description}

\paragraph*{Real-World Dataset}
We construct a graph-based dataset from São Paulo’s street network (Brazil’s largest city, $\sim$12M residents). Crime records from the São Paulo Police Department (1.65M incidents, 2006–2016) are integrated with static socioeconomic and infrastructure data \cite{salinas2022cityhub,hassan2024modeling}. Each crime incident, with its temporal and spatial information, is mapped to the nearest street edge and then assigned to the closest graph node. Infrastructure features include counts of bus, metro, and train stations within 200m (Geosampa), and a binary indicator of proximity ($\leq$500m) to subnormal agglomerates (IBGE). Socioeconomic attributes from the Brazilian Census are aggregated at the census tract level and propagated to nodes within each tract. The variables include: average household and householder income, unemployment rate, literacy (ages 7–15), and population shares for three age groups (under 18, 18–65, over 65).

\paragraph*{Synthetic Dataset}
To enable controlled evaluation, we construct a synthetic dataset that preserves São Paulo's spatial graph structure. Nodes are clustered geographically via \kmeans\ into 12 spatial clusters. Each cluster is assigned 11 static features, sampled from Gaussian distributions with cluster-specific means and equal variance to induce spatial heterogeneity. Dynamic features model monthly crime activity over 144 time steps, with cluster-specific Fourier-based temporal patterns and node-level noise to introduce variability. This ensures both spatial and temporal variability that reflects localized patterns.

\paragraph*{Model Tuning} Models are trained and tuned on both datasets using \glspl{gae} to capture latent spatiotemporal representations.
The number of layers, the dimensionality of the hidden layers, the activation functions, and the dropout rates are selected through a grid search aimed at minimizing the \gls{gae} feature reconstruction loss.
This tuning ensures robust and generalizable performance through systematic experiments.

\section{Fusion Evaluation with synthetic data} \label{sec: synthetic_evaluation}

We design an evaluation framework with synthetic data to systematically analyze how the four fusion-encoding strategies perform the embeddings. Such analysis is performed based on cluster preservation,  quantified using silhouette-based metrics that capture cohesion and separation.



\smallskip

\noindent\emph{Distance Metric:} For data instances $\{x_i\}_{i=1}^N$, we define pairwise distances as $d(x_i, x_j) = \|x_i - x_j\|_2$ for static/fused data, and $d(x_i, x_j) = \mathrm{DTW}(x_i, x_j)$ for time series.

\smallskip

\noindent\emph{Cohesion and Separation:} Intra-cluster cohesion is $a_k = \frac{1}{|C_k| (|C_k| - 1)} \sum_{i \neq j} d(x_i, x_j)$, and separation between clusters $C_k$ and $C_l$ is $b_{kl} = \min d(x_i, x_j)$ for $x_i \in C_k$, $x_j \in C_l$.

\smallskip

\noindent\emph{Dissimilarity and Silhouette:} The cluster's dissimilarity is measured using silhouette score ($S_{kl}$) as:
\[
S_{kl} = \frac{1}{2} \left( \frac{b_{kl} - a_k}{\max(a_k, b_{kl})} + \frac{b_{kl} - a_l}{\max(a_l, b_{kl})} \right),
\]

\noindent Higher $S_{kl}$ values indicate better-separated clusters. The silhouette is not computed in the latent space but rather in the original space as follows: Given the latent representation $z_i$ of each data instance $x_i$, we apply \kmeans\ to group the $z_i$ according to their similarity. Two Silhouette Scores are computed, one for the static and another for the dynamic data, using the cluster's IDs computed in the latent space.
Preservation of the original-space proximities of the samples in the latent space indicates the accuracy of the encoder in capturing data features.



\smallskip

\noindent\emph{Evaluation Pipeline} Each model is evaluated through: (1) \gls{tsne}~\cite{Maaten08} for visualization, (2) \kmeans\ clustering, and (3) quantitative evaluation using Silhouette Scores. Then, we enable fair comparison of fusion strategies in terms of latent-space structure. Fig.~\ref{fig:fusion_models} (middle row) shows that all but one model successfully produce 12 well-separated clusters. Only \textbf{M1 Static} model shows an overlapping pair of clusters due to similarity of their static feature values.


The bottom row in Fig.~\ref{fig:fusion_models} illustrates the clustering quality of embeddings across static, dynamic, and three fusion-based models using Silhouette Scores, where higher values (closer to 1) indicate well-separated clusters and lower values (closer to -1) suggest poor or overlapping clusters. The x-axis shows static Silhouette Scores and the y-axis shows dynamic scores, dividing each plot into four quadrants: the top-right indicates clusters well-formed in both static and dynamic spaces; top-left suggests clusters poorly defined statically but strong dynamically; bottom-right indicates strong static representation but weak dynamic data; and bottom-left reflects poorly formed clusters in both.
This analysis offers a comprehensive understanding of how static and dynamic information are been handled by the encoders in order to generate fused representations. The more concentrated the nodes are in the top-right quadrant, the better the encoder is fusing the data.

In \textbf{M1 Static}, points are mostly concentrated in the top-right and bottom-left quadrants, suggesting that some clusters are consistently well-formed (high static and dynamic dissimilarity), while others remain weak across both modalities. The \textbf{M1 Dynamic} exhibits a strong presence in the top left and right quadrants, indicating that many clusters are poorly formed in the static space but well-separated in the dynamic space. \textbf{M2}, which trains a single \gls{gae} on the concatenated features, shifts more points into the top-right quadrant compared to the individual models. This indicates that combining static and dynamic inputs allows the model to extract mutually reinforcing features. \textbf{M3}, which merges embeddings from independently trained models, is not so effective, concentrating many points in the bottom-left quadrant (poor static and dynamic representations). \textbf{M4}, which adds a third model on top of concatenated embeddings, yields a stable clustering structure, with most points concentrated in the top-right quadrant. 

In summary, \textbf{M4} demonstrates the best clustering quality across both modalities, followed by \textbf{M2}. These results highlight the advantage of deeper integration when combining temporal and static features.





\section{Visualization Assisted Evaluation}
\label{sec:visual-analysis-tool}

This section presents the design and implementation of the visualization tool developed to support the analysis of autoencoder embedding quality.


\vspace*{-0.1cm}
\subsection{Analytical Tasks and System Requirements} 
\label{sec:requirements_and_tasks}
\vspace*{-0.1cm}
Our prior experience in urban data analysis informed both case study design and tool development. In particular, we raised two main requirements to be accounted for when developing the analytical tool:\
\textbf{R1 – Compare Fusion Mechanisms.} Enable comparison of different fusion strategies, especially their impact on locality preservation.\
\textbf{R2 – Understand Node Attributes.} Allow exploration of how original attributes contribute to fusion and interpretation.

We then define key analytical tasks that the visualization tool must support for effective exploration of embedded data.

\noindent\textbf{T1 – Embedding Visualization.} Visualize the different embeddings for overall comparison. This task supports R1.\
\textbf{T2 – Pattern Discovery.} Depict original attributes for focused analysis. This task supports R2 by showing attribute influence on pattern formation.\
\textbf{T3 – Filtering.} Select embedded instances to view locations and detailed attributes (It also supports R2).\
\textbf{T4 – Linked Views.} Highlight selected instances in all views. \textcolor{black}{This task accounts for R1 and R2 by keeping context across embeddings and attributes.}\
\textbf{T5 – Feature Comparison.} Compare selected nodes with the full dataset \textcolor{black}{(It supports to R2).}\
\textbf{T6 – Temporal Analysis.} Visualize temporal evolution of embeddings and attributes. Supports R1 and R2 by showing their evolution over time.

\begin{table}[tb]
\vspace*{-0.5cm}
\caption{Visual components, analytical tasks \textcolor{black}{and requirements}.}
\label{tab:taskvscomponent}
\vspace*{-0.2cm}
\resizebox{0.5\textwidth}{!}{  
\begin{tabu}{
    *{2}{l}
    *{6}{c}
}
\toprule
    & \textbf{Sec.} & \textbf{T1} & \textbf{T2} & \textbf{T3} & \textbf{T4} & \textbf{T5} & \textbf{T6} \\
\midrule
\textbf{Projection View}             & \ref{subsec:V2}& \checkmark & \checkmark  & \checkmark  &  \checkmark          &             &             \\
\textbf{Map View} & \ref{subsec:V3}             &             &             & \checkmark  & \checkmark &             &             \\
\textbf{Discrete Features Bar plots}& \ref{subsec:V4}&             &             & \checkmark  &            & \checkmark  &             \\
\textbf{Features Box plots}& \ref{subsec:V5}&             &             & \checkmark  &            & \checkmark  &             \\ 
\textbf{Time Series Crimes}& \ref{subsec:V6}&             &             & \checkmark  &            &             & \checkmark\\ \hline
\textcolor{black}{\textbf{Requirements Addressed}} & 
\textcolor{black}{--} & 
\textcolor{black}{R1} & 
\textcolor{black}{R2} & 
\textcolor{black}{R2} & 
\textcolor{black}{R1, R2} & 
\textcolor{black}{R2} & 
\textcolor{black}{R1, R2} \\
\bottomrule
\end{tabu}

}
\vspace*{-0.6cm}
\end{table}

\vspace*{-0.1cm}
\subsection{Visual Components}
\vspace*{-0.1cm}
\label{sec:visual_components}

\begin{figure}[!h]
\vspace*{-0.4cm}
\centering
\includegraphics[width=3.5in]{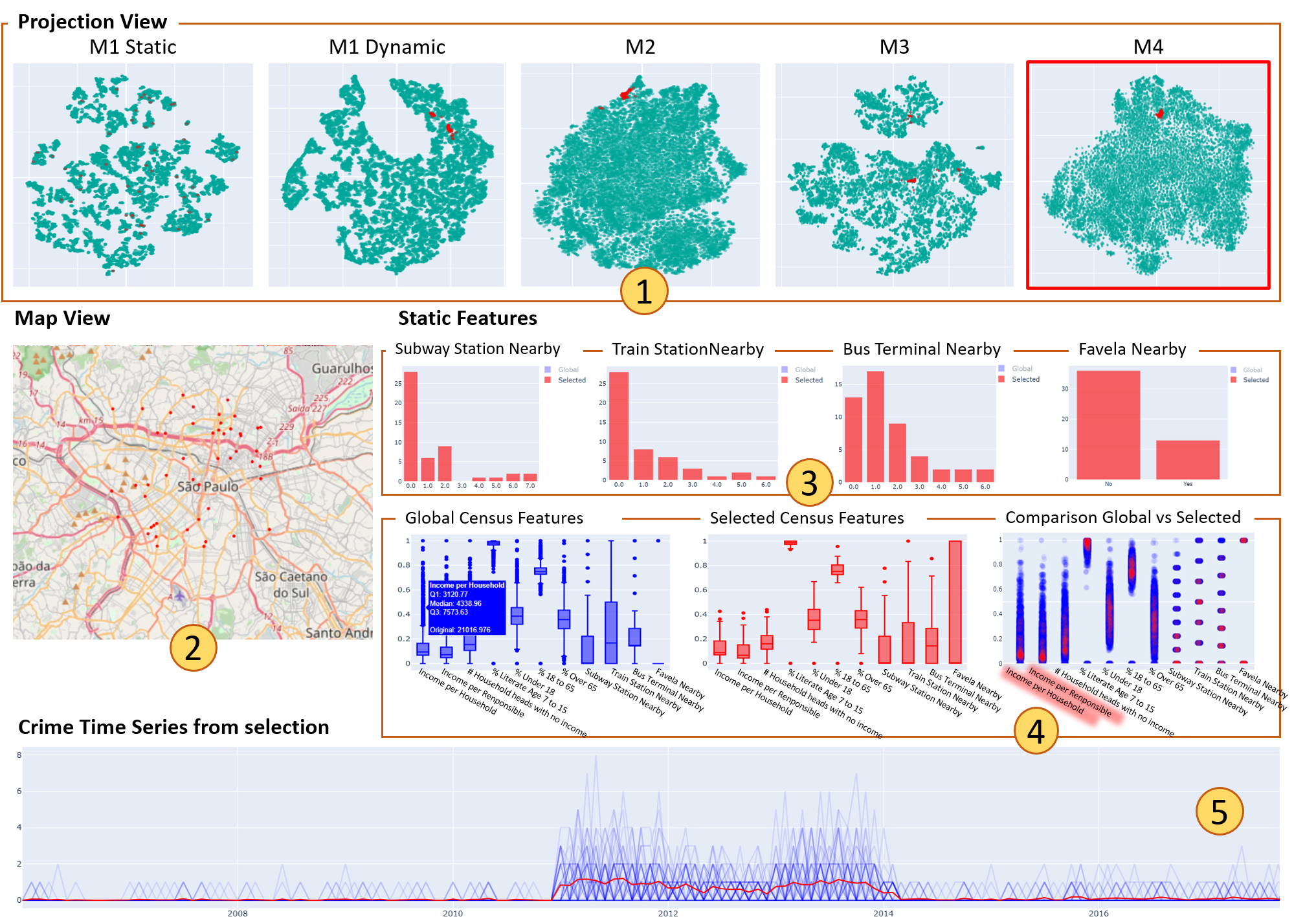}
\vspace*{-0.5cm}
\caption{The interface includes filters for static and dynamic features and five coordinated views: projection, map, bar plots, box plots, and time series. This figure presents \textbf{Case Study IV}, revealing hidden static and dynamic patterns.
}
\label{fig_cityhub}
\vspace{-0.4cm}
\end{figure}

The \gls{va} tool (Fig.\ref{fig_cityhub}) includes five coordinated views to support the designed tasks (Sec.\ref{sec:requirements_and_tasks}). The task–component relations are summarized in Table~\ref{tab:taskvscomponent}.


\subsubsection{Projection View}
\label{subsec:V2}
displays five 2D \gls{tsne} scatter plots of static, dynamic, and fused embeddings (Fig.~\ref{fig_cityhub}.1). Each point represents a street corner with associated static and dynamic data, and lasso selections are synchronized between views for pattern and group analysis.


\subsubsection{Map View}
\label{subsec:V3}
depicts selected instances on map (Fig.~\ref{fig_cityhub}.2).


\subsubsection{Discrete Features Bar Plots}
\label{subsec:V4}
show the frequency distribution of categorical or binned static features (Fig.~\ref{fig_cityhub}.3), comparing global data (blue) with selected subsets (red) to highlight feature-level patterns.


\subsubsection{Feature Distribution Plots}
\label{subsec:V5}
visualize static features across three coordinated views (Fig.~\ref{fig_cityhub}.4) using min-max normalization, with original values shown on hover. The left boxplot shows the full dataset, the middle one shows the selected nodes, and the right plot shows dispersion diagrams overlaying global and selected nodes to highlight differences.

\subsubsection{Time Series}
\label{subsec:V6}

displays crime time series for the selected nodes (Fig.~\ref{fig_cityhub}.5), with each line-plot corresponding to a node. The average trend across all time series is shown in red.

\section{Case Studies}
\label{sec:case_studies}

    To demonstrate the value of our analytical tool, we present case studies that highlight how different fusion models encode spatiotemporal patterns rather than specific urban regions.

\begin{figure}[!t]
\centering
\vspace*{-0.7cm}
\includegraphics[width=3.5in]{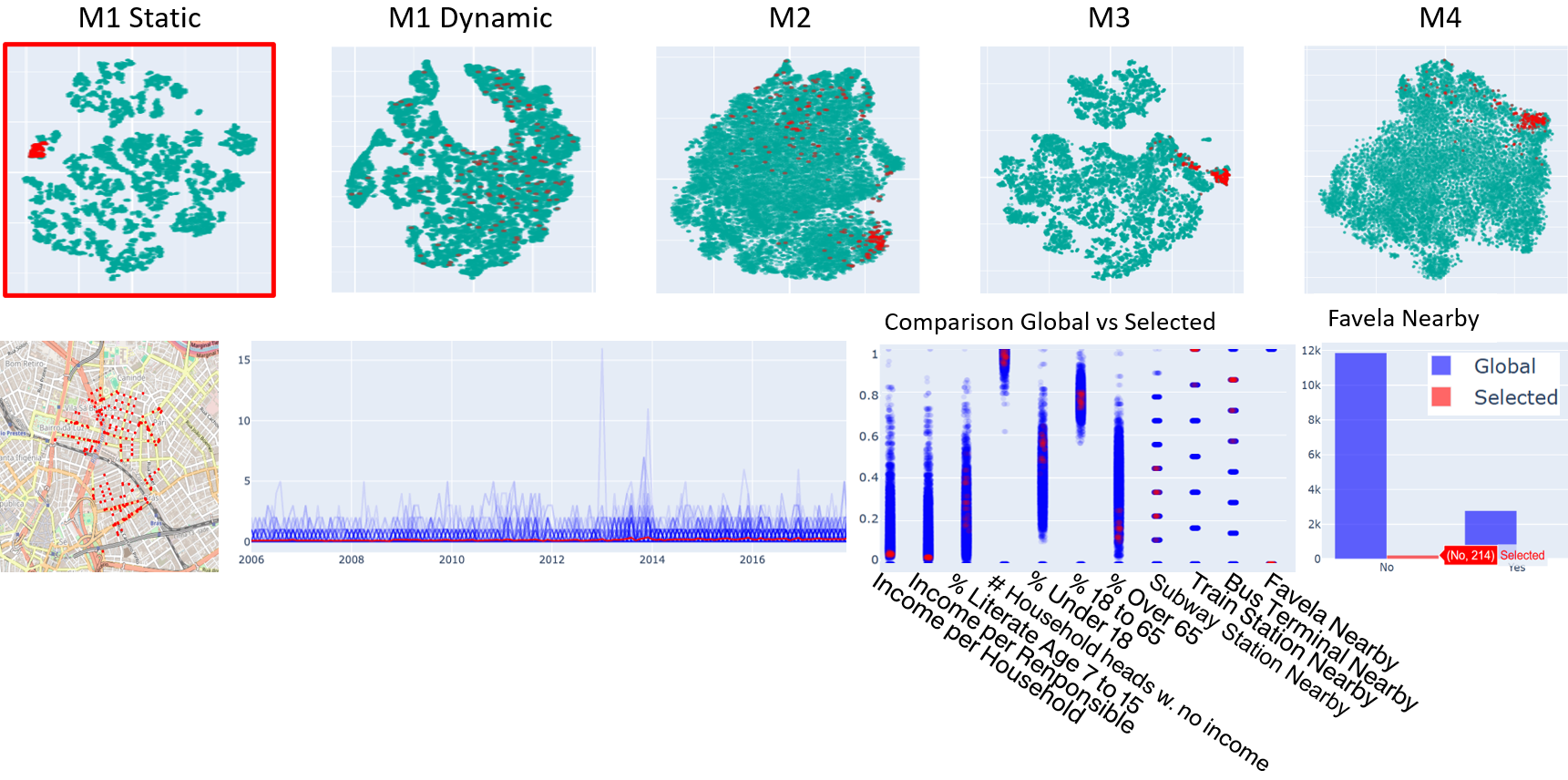}
\vspace*{-0.5cm}
\caption{\textbf{Case Study I.} 
Selection in \textbf{M1 Static} is preserved in \textbf{M2}, \textbf{M3}, and \textbf{M4}; this cluster shows geographic proximity and high socioeconomic values.
}
\label{fig_cs1}
\vspace*{-0.6cm}
\end{figure}

\noindent \textbf{Case Study I.} 
In Fig.~\ref{fig_cs1}, we observe that the subset selected on the \textbf{M1 Static} shows up relatively concentrated across fusion models. Particularly, in \textbf{M3} and \textbf{M4}, suggesting that these fused models are capable of capturing underlying static patterns. 
This cluster appears to be shaped by socioeconomic characteristics, as it shows lower values for both Income per Household and Income per Responsible when compared to the overall distribution (see the red dispersion diagram, third from the bottom-right). Another notable aspect is that the selected nodes are not located near favelas (see the bar plot at the bottom), which is also confirmed in the Map View, where a spatial concentration is visible. While the cluster is well-defined in \textbf{M1 Static}, it becomes spread in \textbf{M1 Dynamic}, which may explain the lack of a clear trend in the time series view.

\begin{figure}[!t]
\centering
\vspace*{-0.8cm}
\includegraphics[width=3.5in]{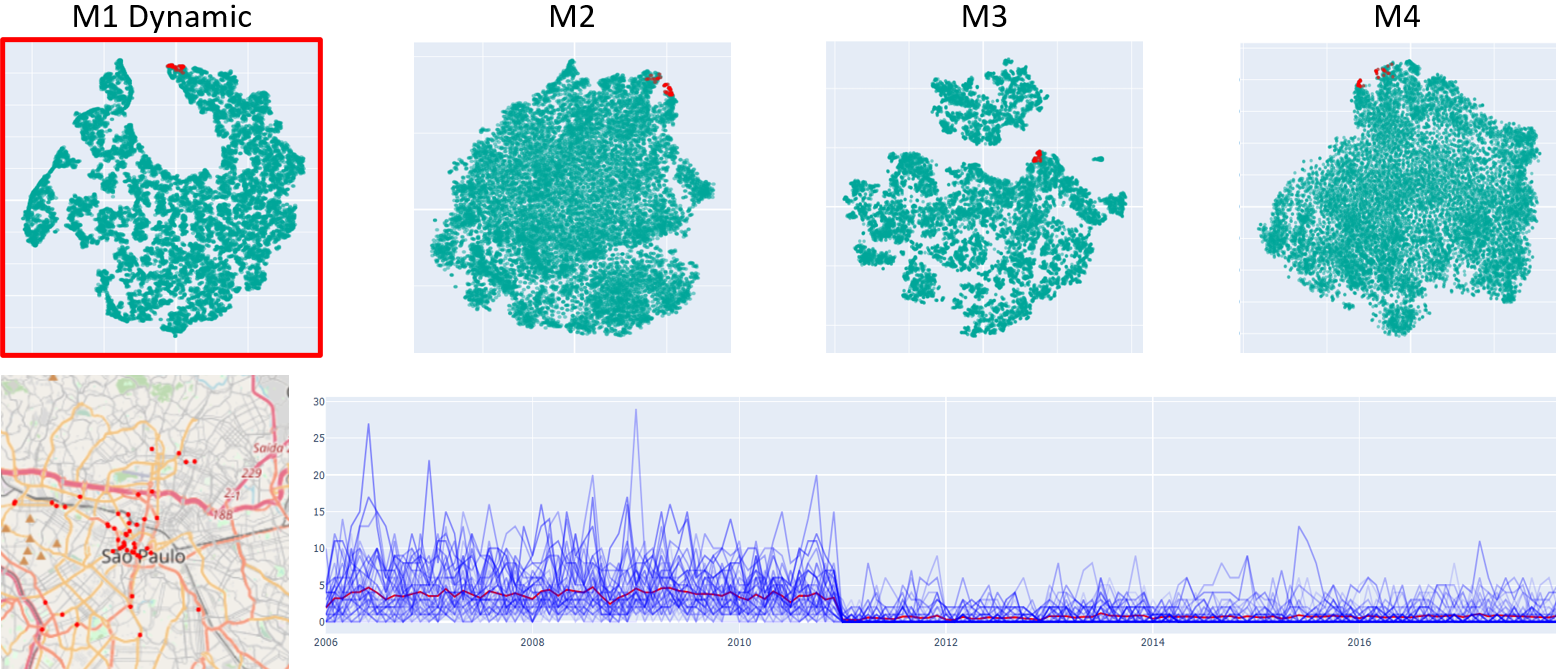}
\vspace*{-0.6cm}
\caption{\textbf{Case Study II.} 
The selection in the \textbf{M1 Dynamic} projection is preserved across the fused models (\textbf{M2}–\textbf{M4}). The corresponding time series display elevated crime levels from 2006 to 2011.
}
\label{fig_cs2}
\vspace*{-0.4cm}
\end{figure}

\begin{figure}[!t]
\centering
\includegraphics[width=\linewidth]{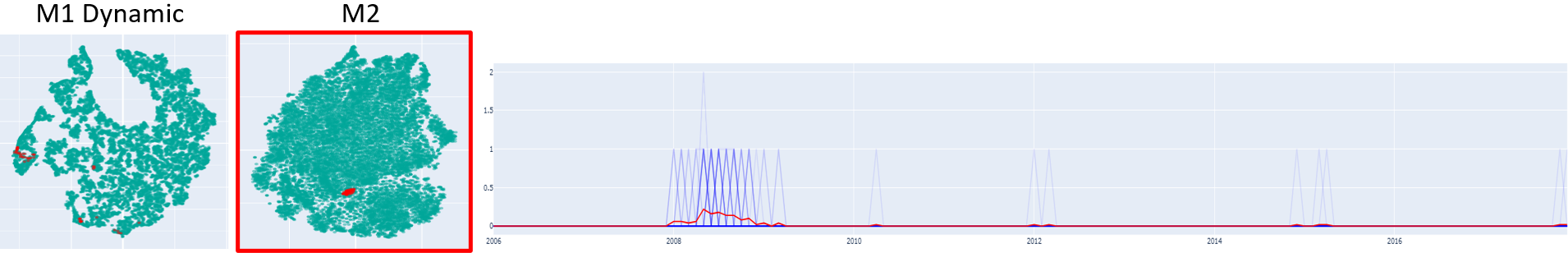}
\vspace*{-0.5cm}
\caption{\textbf{Case Study II.} \textbf{M2} reveals temporal patterns undetected in pure dynamic projections.
}
\label{fig_cs3}
\vspace*{-0.7cm}
\end{figure}

\noindent \textbf{Case Study II.}
Fig.~\ref{fig_cs2} depicts nodes selected from the \textbf{M1 Dynamic} scatter plot. Notice that the time series associated with the selected nodes exhibits a well-defined pattern (bottom right plot): they consistently show high crime levels from 2006 to 2011, with frequent spikes and fluctuations. In early 2011, there is a sharp drop in reported crime, after which the values stabilize at much lower levels until the end of 2017.
Therefore, those nodes bear a similar crime-related pattern, transitioning from a high-crime to a low-crime period. 
The map view shows that most nodes cluster in central São Paulo, with some distant locations showing similar behavior.
Notice that the selected nodes are also tightly grouped in the fused models (\textbf{M2}, \textbf{M3}, and \textbf{M4}), showing that the \gls{gae} fusing models are properly handling dynamic data, even more stringently than static data. Fig.~\ref{fig_cs3} further corroborates this fact, where a subset of nodes is selected in the scatter plot \textbf{M2}. 
The time series associated with the selected nodes has a well-defined pattern, with crimes concentrated in 2008.
Interestingly, the selected group is dispersed in the \textbf{M1 Dynamic} layout, indicating that the fusion mechanism captures crime patterns more effectively than the dynamic-only model.


\begin{figure}[h]
\vspace*{-0.5cm}
\centering
\includegraphics[width=\linewidth]{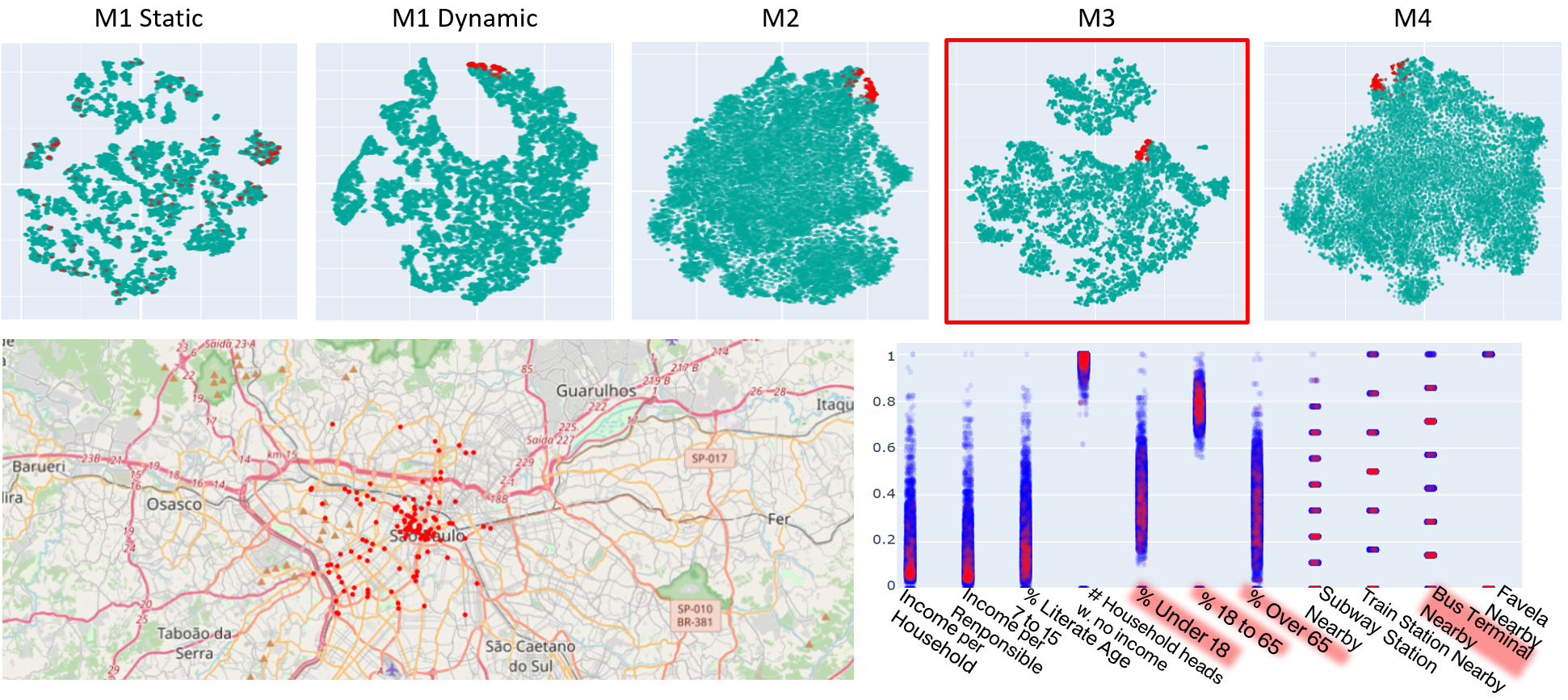}
\vspace*{-0.5cm}
\caption{\textbf{Case Study III.} Group selected in \textbf{M3} projection layout. Fusing models seems to give more attention to dynamic than static features when generating the embeddings.  
}
\vspace*{-0.3cm}
\label{fig_cs4}
\end{figure}

\noindent \textbf{Case Study III.} 
The selection in the \textbf{M3} layout is preserved in all models except \textbf{M1 Static} (Fig.~\ref{fig_cs4}), where nodes are significantly spread.
This result reinforces that the fusion models are accounting for the dynamic feature more than in the static ones.
However, we can infer that static features were indeed considered, as several attributes differ from the global distribution in the dispersion diagram, particularly those related to age. The selected group is characterized by a lower proportion of children and elderly individuals, and a higher concentration of adults aged 18 to 65.

\noindent \textbf{Case Study IV.} The group selected in \textbf{M4} is not strongly preserved in the other models (see Fig. ~\ref{fig_cityhub}). However, a clear pattern emerges in the time series, with a higher concentration observed from early 2012 to early 2014, a trend that was not captured by the \textbf{M1 Dynamic} model.
Although \textbf{M1 Static} does not preserve the cluster either, the boxplots reveal that the selected group has a particular pattern of static feature, showing the \textbf{M4} could better capture both static and dynamic patterns simultaneously. For instance, the features Income per Household and Income per Responsible, with values notably low, as indicated by the red dots in the dispersion diagram. Moreover, the high percentage of Literate Age 7 to 15 suggests these static features were effectively considered by the fusion mechanism of \textbf{M4}.
Regarding transportation, the selected nodes tends to be close to a reduced number of subway stations, train stations, and bus terminals. This may be attributed to the tendency of these neighborhoods to rely on private modes of transportation. 

\section{Discussion}


The experiment with synthetic data demonstrates that the fusion models effectively integrate static and dynamic information, with model \textbf{M4}, which features a two-stage fusion scheme, showing particularly strong performance. The visualization tool proved instrumental in revealing how fusion is being performed, indicating that the models tend to emphasize dynamic features while still accounting for static ones. This balance allows for the identification of locations with similar static and dynamic patterns, making it easier to get insights from complex multimodal data.

In essence, the visualization tool enhances users' confidence in the quality and reliability of the embeddings, while also providing a means to compare different data fusion models. These findings highlight the value of visualization in analyzing fusion strategies, an aspect largely overlooked in the existing literature \cite{baltruvsaitis2018multimodal}. Thus, this work makes a significant contribution by emphasizing the need for visualization-assisted tools in the evaluation and understanding of data fusion techniques. In fact, it represents a first step toward establishing visualization as a fundamental resource in the analysis of data fusion models. 
\section{Conclusion}
{In this work, we evaluated several \gls{gae}-based fusion models for the joint analysis of static and dynamic features in urban analytics. We developed a \gls{va} system to investigate models' performance. Through a series of case studies, we demonstrated that the fused latent representations effectively capture heterogeneous data patterns, enabling meaningful interpretation. Future work may explore multi-resolution spatial aggregation and automated pattern detection.}

%

\vspace*{-0.06cm}
\section*{Acknowledgment }
\vspace*{-0.12cm}
This work was supported by FAPESP (\#2020/07012-8, \#2022/09091-8, \#2023/16334-7), CNPq (\#307184/2021-8), CAPES and by the Deutsche Forschungsgemeinschaft (\#360330772). The opinions, hypotheses, conclusions, and recommendations expressed in this material are the responsibility of the authors and do not necessarily reflect the views of FAPESP, and CNPq, and CAPES.
\vspace*{-0.17cm}



\bibliographystyle{IEEEtran}
\bibliography{example}
%
%


\end{document}